\documentclass[letterpaper]{article} 
\usepackage{aaai2026}  
\usepackage{array}   
\usepackage{times}  
\usepackage{helvet}  
\usepackage{courier}  
\usepackage[hyphens]{url}  
\usepackage{graphicx} 
\usepackage[utf8]{inputenc}
\usepackage{amssymb}   
\usepackage{pifont}    
\usepackage{subcaption} 
\usepackage{multirow}
\usepackage{pdflscape} 
\usepackage{adjustbox} 
\usepackage{booktabs} 
  
\usepackage{natbib}  
\usepackage{caption} 
\usepackage{algorithm}
\usepackage{algorithmic}
\usepackage{newfloat}
\usepackage{listings}
\DeclareCaptionStyle{ruled}{labelfont=normalfont,labelsep=colon,strut=off}
\floatstyle{ruled}
\newfloat{listing}{tb}{lst}{}
\floatname{listing}{Listing}

\title{C-DGPA: Class-Centric Dual-Alignment Generative Prompt Adaptation}
\author{
  Chao Li\textsuperscript{1}, 
  Dasha Hu\textsuperscript{1}\thanks{*Corresponding author.}, 
  Chengyang Li\textsuperscript{1}, 
  Yuming Jiang\textsuperscript{1}, 
  Yuncheng Shen\textsuperscript{2}
}

\affiliations{
  $^{1}$ College of Computer Science, Sichuan University, Chengdu, China\\
  $^{2}$ College of Physics and Information Engineering, Zhaotong University, Zhaotong, China
}

\begin{document}

\maketitle

\begin{abstract}
Unsupervised Domain Adaptation transfers knowledge from a labeled source domain to an unlabeled target domain. Directly deploying Vision-Language Models (VLMs) with prompt tuning in downstream UDA tasks faces the significant challenge of mitigating domain discrepancies. Existing prompt-tuning strategies primarily align marginal distribution, but neglect conditional distribution discrepancies, leading to critical issues such as class prototype misalignment and degraded semantic discriminability. To address these limitations, the work proposes C-DGPA: Class-Centric Dual-Alignment Generative Prompt Adaptation. C-DGPA synergistically optimizes marginal distribution alignment and conditional distribution alignment through a novel dual-branch architecture. The marginal distribution alignment branch employs a dynamic adversarial training framework to bridge marginal distribution discrepancies. Simultaneously, the conditional distribution alignment branch introduces a Class Mapping Mechanism (CMM) to align conditional distribution discrepancies by standardizing semantic prompt understanding and preventing source domain over-reliance. This dual alignment strategy effectively integrates domain knowledge into prompt learning via synergistic optimization, ensuring domain-invariant and semantically discriminative representations. Extensive experiments on OfficeHome, Office31, and VisDA-2017 validate the superiority of C-DGPA. It achieves new state-of-the-art results on all benchmarks.

\end{abstract}

\begin{links}\link{Code}{https://anonymous.4open.science/r/C-DGPA-B37F}
\end{links}

\section{Instruction  }

Unsupervised Domain Adaptation (UDA) transfers knowledge from a labeled source domain to an unlabeled target domain to enhance learning effectiveness. The core challenge lies in reducing distribution discrepancies between domains to improve model generalization on unlabeled target data without sacrificing semantic discriminability\cite{wilson2020survey,zhu2023generalized}. Traditional UDA methods rely on adversarial training\cite{ganin2015unsupervised} and metric learning\cite{saito2018maximum,tang2020unsupervised,zhang2020knowledge} to narrow distribution gaps and achieve domain alignment. Although partially effective, these methods often incur semantic information loss and fail to disentangle domain knowledge from semantic content effectively\cite{zhang2022domain,tang2020unsupervised}.

Concurrently, large-scale visual language models (VLMs), particularly CLIP\cite{radford2021learning}, exhibit remarkable performance in UDA tasks due to their strong image-text integration capabilities\cite{hu2024reclip,lai2024empowering}. VLMs map visual and textual representations into a shared embedding space, reducing domain differences, and improving discriminative performance. Prompt tuning has emerged as an effective strategy for adapting VLMs to downstream unlabeled data. State-of-the-art techniques like CoOp\cite{zhou2022learning} and MaPLe\cite{khattak2023maple} deliver strong performance on specific tasks: CoOp learns soft text prompts, while MaPLe employs visual-linguistic prompts to enhance interaction. Despite mitigating domain discrepancies, these techniques focus primarily on aligning marginal probability distribution through prompt, overlooking cross-domain conditional distribution discrepancies. This oversight causes:

• Class Prototype Shift\cite{tang2020unsupervised} : Feature space misalignment between source and target domain prototypes (distribution centers) for the same class.

• Semantic Discriminability Degradation: Blurred class boundaries persist even after global feature alignment, leading to target sample misclassification.

Long\cite{long2015learning} demonstrated that the total domain discrepancy decomposes into the sum of marginal and conditional distribution discrepancies. Consequently, conditional alignment can significantly improve prompt performance against domain discrepancies, enabling deeper understanding and adaptation to diverse data distributions.

Inspired by Long et al, the work proposes C-DGPA (Class-Centric Dual-Alignment Generative Prompt Adaptation), the first class-centric dual-alignment method for generative prompt adaptation. C-DGPA employs dual branches (marginal distribution  and conditional distribution alignment branch) to address limitations in cross-domain learning. The Marginal Distribution Alignment Branch employs adversarial training to minimize marginal distribution discrepancies, generating domain-invariant features through a dynamic minimax optimization framework with Gradient Reversal Layer (GRL). Meanwhile, the Conditional Distribution Alignment Branch addresses conditional distribution discrepancies via the proposed Class Mapping Mechanism (CMM), which standardizes semantic prompt representations for specific classes by aligning cross-domain class prototypes. This dual-branch architecture synergistically optimizes prompt parameters: the Marginal Distribution Alignment Branch ensures feature invariance across domains, while the Conditional Distribution Alignment Branch mitigates source domain over-reliance and enhances target domain discriminability by refining class-sensitive semantics. Key innovations include:

• The work proposes C-DGPA, a novel Class-Centric Dual-Alignment framework for generative prompt adaptation in UDA. C-DGPA is the first to integrate the Marginal Alignment Distribution Branch and Conditional Distribution Alignment Branch within a unified prompt tuning paradigm via synergistic dual-branch optimization.

• The work develops a dynamic adversarial alignment framework within the Marginal Distribution Alignment Branch. This framework, incorporating a minimax objective and a GRL, effectively aligns the marginal distribution across domains, generating domain-invariant prompt features crucial for subsequent class alignment.

• The work introduces the CMM as the core of the Conditional Distribution Alignment Branch. CMM explicitly addresses the critical challenge of conditional distribution discrepancies by mapping prompt features to a domain-invariant class prototype space, supervised by the conditional alignment loss. This mechanism is key to mitigating class prototype shift and enhancing semantic discriminability.

Our research findings show that C-DGPA performs excellently on benchmark datasets, including OfficeHome, Office31, and VisDA-2017. Compared with existing state-of-the-art methods, C-DGPA achieves significant performance improvements, proving its effectiveness in unsupervised domain adaptation tasks.

\section{Related Work}
\subsection{Unsupervised Domain Adaptation}
The core objective of UDA is to learn domain-invariant features by aligning distributions across domains. Early methods rely on statistical metrics such as maximum mean discrepancy (MMD)\cite{long2015learning} to minimize feature distribution gaps. Adversarial learning further advanced UDA by training domain-invariant features through minimax optimization. Domain-Adversarial Neural Networks (DANN)\cite{ganin2016domain} introduced gradient reversal layers to align domains, while Smooth Domain-Adversarial Training (SDAT)\cite{rangwani2022closer} improved generalization by smoothing source loss.
\subsection{Vision-Language Models (VLMs)}
Large-scale VLMs like CLIP\cite{hu2024reclip,lai2024empowering} enable cross-modal alignment through contrastive pre-training, offering strong zero-shot transferability. To adapt VLMs to downstream tasks, prompt tuning has emerged as a key paradigm. CoOp replaced manual prompts with learnable vectors, and MaPLe extended this to multimodal prompts. PDA\cite{bai2024prompt} introduced dual-branch alignment for further marginal alignment distributions. However, existing methods primarily align marginal distributions, neglecting conditional distribution discrepancies. C-DGPA addresses these gaps by integrating a dynamic adversarial alignment framework for marginal distribution with CMM for conditional distribution.
\subsection{Visual prompt tuning}
By optimizing the input prompt vectors, the model can better understand and execute specific tasks. This approach is particularly prominent in few-shot\cite{gu2021ppt} and zero-shot\cite{reynolds2021prompt} learning environments. In recent years, researchers have proposed Visual Prompt Tuning (VPT)\cite{jia2022visual,zhu2023bridging,pfeiffer2023modular}, which applies prompts to the visual domain. The core idea of prompting is to embed new prompt vectors for parameter training on the basis of the original model, while freezing the parameters of the original model. This allows the model to update only the parameters of the prompt during training, thereby achieving the desired optimization effect.
\section{Preliminaries}
\subsection{Unsupervised Domain Adaptation}
Unsupervised Domain Adaptation (UDA) focuses on transferring knowledge from a labeled source domain to an unlabeled target domain. The source domain $C_{s}$ comprises a feature space $x_{s}$ and a marginal distribution $P_{\mathrm{s}}(x_{s})$, while the target domain $C_t$ consists of a feature space $x_t$ and a marginal distribution $P_t(x_t).$ In the source domain, there is a sample set $x_s=\{x_1^s,x_2^s,...,x_N^s\}$ with corresponding labels $y_{s}=\{y_{1}^{s},y_{2}^{s},...,y_{N}^{s}\}.$ In contrast, the target domain only contains a sample set $x_t=\{x_1^t,x_2^t,...,x_M^t\}$ , where N and M represent the number of samples in the source and target domains, respectively. A model trained on the source domain is intended for testing on the target domain. However, in many application scenarios, such as face recognition, various factors(e.g., lighting, angles, expressions) can affect the target domain's data distribution. UDA addresses these challenges by mitigating the impact of such factors and enabling effective knowledge transfer from the labeled source domain to the unlabeled target domain.
\subsection{Contrastive Language-Image Pretraining}
CLIP (Contrastive Language-Image Pretraining) is a model capable of jointly processing images and text. It learns the relationship between images and text through pretraining. In downstream tasks, the pretrained CLIP model is adapted to specific tasks via manually designed prompts. The matching score between an image and text is calculated based on the cosine similarity $\langle\widehat{w}_{i},z\rangle$ between the image representation z and the text representation corresponding to the i-th class. The image representation is obtained by processing the input image with an image encoder, while the text representation is extracted from a text encoder using a prompt description associated with the i-th class. The probability that an image belongs to the i-th class is given by:
\begin{equation}
    P(y=i|\mathrm{x}) = \frac{\exp(\langle\widehat w_i,\mathrm{z}\rangle/\tau)}{\sum_{j=1}^K\exp\left(\langle\widehat w_j,\mathrm{z}\rangle/\tau\right)}
    \label{eq:softmax}
\end{equation}
\subsection{Feature Bank Construction}
Confidence scores are generated for source domain images. Target domain pseudo-labels are then determined. Subsequently, the top C features per class from both domains form a K-class feature bank (C per class). Class center features establish the source bank $f_{s} \in \mathbb{R}^{d}$ and target bank $f_{t} \in \mathbb{R}^{d}$. This ensures feature bank robustness and significantly improves cross-domain learning performance.
\begin{figure}
    \centering
    \includegraphics[width=1\linewidth]{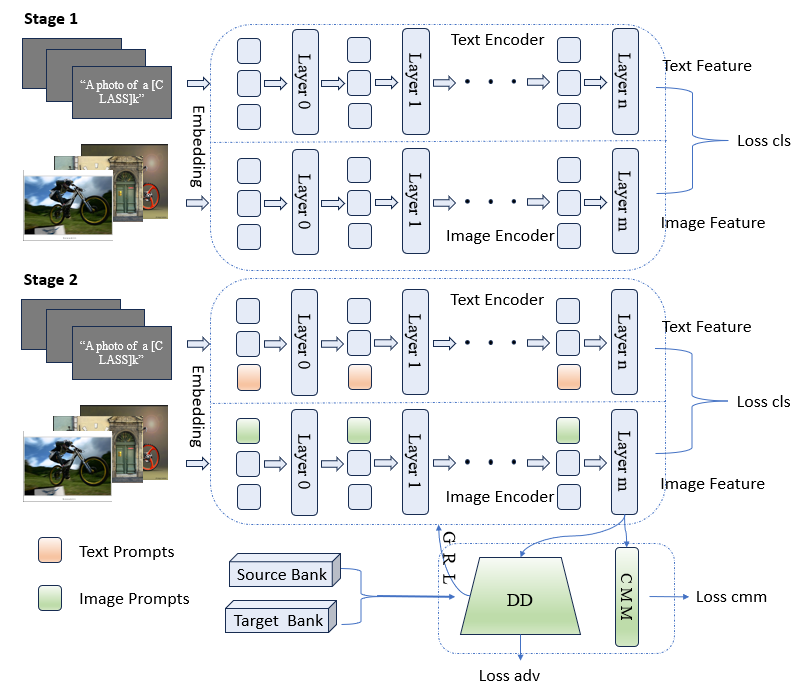}
    \caption{The architecture of the C-DGPA model. This figure illustrates the structure of our model at different stages. The Source Bank and Target Bank represent the feature banks of the source and target domains, respectively. CMM is used to map the image features I to the class space, while DD denotes the domain discriminator that generates domain labels based on I. In Stage 2, the blue color indicates the frozen parts, while other colors indicate the parts that need to be trained.}
    \label{fig:enter-label}
\end{figure}
\subsection{Theoretical Foundation for Dual-Alignment}
Unsupervised Domain Adaptation aims to minimize the total domain discrepancies $d_J$ between source and target distributions. As formalized by Long\cite{long2015learning}, the distributions decompose into the sum of marginal distribution discrepancies $\bar{d}_{\mathcal{H}}(P_{s}(x_{s}),P_{t}(x_{t}))$ and conditional distribution discrepancies $d_{\mathrm{C}}(P_{\mathrm{s}}(y_{\mathrm{s}}|x_{\mathrm{s}}),P_{t}(\hat{y}_{t}|x_{\mathrm{t}}))$:
\begin{equation}
    d_{J}(P_{s},P_{t}) = d_{\mathcal{H}}(P_{s},P_{t}) + d_{\mathcal{C}}(P_{s},P_{t})
    \label{eq:my_equation}
\end{equation}
\begin{equation}
    d_{C}(P_{s},P_{t}) = \mathbb{E}[\parallel\mathbb{P}(y_{s}|x_{s}) - \mathbb{P}(\hat{y}_{t}|x_{t})\parallel]
    \label{eq:distance_expectation}
\end{equation}
Here, $\hat{y}_{t}$ denotes the pseudo-labels generated for the target domain\cite{lai2023padclip,litrico2023guiding}. Critically, the Ben-David theorem\cite{ben2010theory} establishes
that the target domain error $\epsilon_T(p)$ is bounded by
\begin{equation}
    \epsilon_{T}(p) \leq \epsilon_{S}(p) + d_{J}(P_{S}, P_{T}) + \lambda
    \label{eq:domain_adaptation_bound}
\end{equation}
where $\lambda$ represents the optimal joint error. This theoretical
decomposition necessitates a dual-alignment strategy:

• Marginal Distribution Alignment Branch resolves marginal distribution discrepancies \begin{equation}
d_{\mathcal{H}}(P_{s}(x_{s}), P_{t}(x_{t})) \to 0
\end{equation}

• Conditional Distribution Alignment Branch resolves conditional distribution discrepancies\begin{equation}
d_{\mathcal{C}}(P_s(y_s|x_s), P_t(\hat{y}_t|x_t)) \to 0
\end{equation}
Where x denotes input images, and later alignment is performed on prompt-generated features $I=\bar{f}(x;\theta_{p}).$ Therefore, our proposed dual-branch architecture is explicitly designed to minimize both terms concurrently: The Marginal Distribution Alignment Branch reduces marginal distribution discrepancies $d_{\mathcal{H}}(P_{s}(I_{s}),P_{t}(I_{t}))$ via a Dynamic Adversarial Alignment Framework. The Conditional Distribution Alignment Branch minimizes conditional discrepancies $d_{\mathrm{C}}(P_{\mathrm{s}}(y_{\mathrm{s}}|I_{\mathrm{s}}),P_{t}(\hat{y}_{\mathrm{t}}|I_{\mathrm{t}}))$ through the CMM by projecting $I$ into a semantically invariant space. By synergistically optimizing both objectives, C-DGPA directly suppresses $\bar{d}_{J}(P_{s},P_{t})$,hereby lowering the upper bound of $\epsilon_T(p)-$a theoretical advance unrealized by prior prompt-based UDA methods.
\section{Method}
The architecture of the original CLIP model is depicted in Stage 1 of Figure 1. In this setup, the text encoder produces text features $T=\{T_{1},T_{2},\ldots,T_{n}\}$, while the image encoder generates image features $I=\{I_{1},I_{2},\ldots,I_{n}\}.$ C-DGPA employs a synergistic dual-branch architecture to achieve comprehensive domain adaptation through joint optimization of marginal and conditional distribution alignment(Long et al. 2015). As illustrated in Figure 1(Stage 2), the image features I generated by the prompt are processed concurrently by two distinct branches:The Marginal Distribution Alignment Branch: Imple ments our Dynamic Adversarial Alignment Framework to minimize marginal distribution discrepancy via adversarial training; The Conditional Distribution Alignment Branch: Leverages the Class Mapping Mechanism (CMM) to minimize
conditional distribution discrepancies.
The prompt parameters $\theta_p$ are jointly optimized by the
combined loss $\mathcal{L} = \mathcal{L} _{\mathrm{cls}}+ \gamma _{\mathrm{cal}}\mathcal{L} _{\mathrm{cal}}+ \gamma _{\mathrm{mal}}\mathcal{L} _{\mathrm{mal}}$, enabling
the prompt to progressively learn both domain-invariant
and class-discriminative representations. After fine-tuning the dual branches can be discarded, resulting in a lightweight adapted model. The symbols and their corresponding meanings used in this paper are presented in Tabel 1.
\begin{table*}[htbp]
\centering
\small  
\setlength{\tabcolsep}{3pt}  
\begin{tabular*}{\textwidth}{@{}cc|cc|cc@{}}
\toprule
\textbf{Symbol} & \textbf{Description} & \textbf{Symbol} & \textbf{Description} & \textbf{Symbol} & \textbf{Description} \\
\midrule
$C_s, C_t$ & Source/target domain space & $x_s, x_t$ & Source/target sample set &  $y_s$ & Source labels\\
$L_{\mathrm{adv}}$ & Marginal alignment loss  & $I_s, I_t$ & Source/target image features & $\mathbb{R}^d$ & $d$-dim real space\\
$L_d$ & Domain classifier loss & $d^s, d^t$ & Source/target labels & $\hat{y}_t$ & Target pseudo labels \\
$\gamma_{\mathrm{mal}}$ & Marginal alignment weight & $N, M$ & Source/target sample size & $\mathcal{H}$ & Hypothesis space \\
$\theta_d$ & Domain classifier parameters & $\nabla_{\theta_p}$ & Prompt gradient direction & $\theta_p$ & Prompt parameters \\
$\gamma_{\mathrm{cal}}$ & Conditional alignment weight & $f(x;\theta_p)$ & Image encoder (e.g., CLIP) & $T, I$ & Text/image features  \\
$d_{\mathcal{H}}(P_s, P_t)$ & Marginal distribution discrepancy  & $d_{\mathcal{J}}(P_s, P_t)$ & Total domain discrepancy & $\mathbb{E}$ & Expectation (mean)\\
$\lambda$ & Optimal joint error & $\epsilon_s(p), \epsilon_t(p)$ & Source/target error & $\tau$ & Temperature coefficient \\
$f_s, f_t$ & Source/target feature bank & $I'$ & Augmented image features & $P_{\mathrm{class}}$ & Category prototypes \\
$L_{\mathrm{cal}}$ & Conditional alignment loss & $P_s(y_s|I), P_t(\hat{y}_t|I)$ & Conditional distributions & $P_s(x), P_t(x)$ & Marginal distributions \\
\bottomrule
\end{tabular*}
\caption{Symbol Description }
\label{tab:symbol_description_full}
\end{table*}
\subsection{Marginal Distribution Alignment Branch: Dynamic Adversarial Alignment for Marginal Distribution}
The primary objective of this branch is to align the marginal distribution $P_{s}(I_{s}) \approx P_{t}(I_{t})$ by suppressing domain-specific information. The work achieves this through a dynamically optimized adversarial training framework, formulated as a minimax game combined with a Gradient Reversal Layer (GRL).
\subsubsection{Minimax Game Framework for Adversarial Training:}Based on the ımage teatures $I$ generated by the prompt, adversarial training aims to align the marginal distribution by maximizing the domain discriminator parameters $\theta_d$ and minimizing the prompt parameters $\theta_p.$ Specifically, the marginal alignment loss $\mathcal{L}_\mathrm{mal}$ is defined as:
\[
\mathcal{L}_{mal} = \min_{\theta_p} \max_{\theta_d} \mathcal{L}_d =
\]
\begin{equation}
\min_{\theta_p} \max_{\theta_d} \left( \mathbb{E}_{x_s \sim P_s} [\log C_s(I_s)] + \mathbb{E}_{x_t \sim P_t} [\log(1 - C_t(I_t))] \right)
\label{eq:minmax_loss}
\end{equation}
Here, the domain discriminator seeks to maximize $\theta_d$ to distinguish the origins of features (source vs. target), while the prompt aims to minimize $\theta_p$ to generate domain-invariant features that confuse the discriminator. This process balances the feature distribution, enabling the prompt-generated I to be domain-invariant. The domain discriminator loss $\mathcal{L}_{\mathrm{d}}$ is formulated as:
\begin{equation}
\mathcal{L}_{\mathrm{d}} = -\frac{1}{N}\sum_{i=1}^{N}\log\left(P(\mathrm{d}^{s}|I_{i}^{s})\right) -\frac{1}{M}\sum_{j=1}^{M}\log\left(P(\mathrm{d}^{t}|I_{j}^{t})\right)
\label{eq:loss_d}
\end{equation}
Where $I_{i}^{s}$and $I_{i}^{t}$ denote the image features of the i-th and j-th samples from the source and target domains, respectively.
\subsubsection{Gradient Reversal Layer (GRL) Mechanism:}To minimize $\theta_p$ and facilitate adversarial training through gradient backpropagation, the work incorporates the Gradient Reversal Layer (GRL). During the forward pass, the GRL
maintains an identity mapping. However, during the backward pass, it reverses the gradient of $\mathcal{L}_\mathrm{d}:$
\begin{equation}
\frac{\partial\mathcal{L}_{\mathrm{mal}}}{\partial\theta_p} \leftarrow -\frac{\partial\mathcal{L}_{\mathrm{d}}}{\partial\theta_p}
\label{eq:gradient_update}
\end{equation}
Through gradient reversal, the GRL mechanism forces the prompt to generate semantically rich and domain-agnostic features, achieving marginal distribution alignment. This mechanism updates the prompt parameters as follows:
\begin{equation}
\theta_p \leftarrow \theta_p - \eta \left( \nabla_{\theta_p} L_{cls} + \gamma_{mal} \nabla_{\theta_p} L_{mal} \right)
\label{eq:theta_update}
\end{equation}
This Dynamic Adversarial Alignment Framework ensures the prompt-generated features $I=f(x;\theta_p)$ possess Domain invariance feature $(P_s(I_s)\approx P_t(I_t))$ while retaining Semantic Richness for class discrimination, forming a solid foundation for the Conditional Distribution Alignment Branch. 
\subsection{Conditional Distribution Alignment Branch: Conditional Distribution Alignment via Class Mapping Mechanism}
While marginal alignment is necessary, it is insufficient due to persistent conditional distribution discrepancies $( \mathrm{P} _{\mathrm{s} }( y_{\mathrm{s} }|$ $I_{\mathrm{s} }) \neq \mathrm{P} _{\mathrm{t} }( \hat{y} _{\mathrm{t} }|$ $I_{\mathrm{t} }) )$,manifesting as class prototype shift (Tang et al., 2020). The Conditional Distribution Alignment Branch tackles this by explicitly aligning conditional distribution using the proposed Class Mapping
Mechanism (CMM). CMM operates on the domain-invariant features I from the prompt. Assuming the source domain feature bank $f_s\in\mathbb{R}^d$ and the target domain feature bank $f_{t}\in\mathbb{R}^{d}$ are established as follows:
The work proposes a CMM that maps image features I to the
class space. Specifically, using image features I as the
query $(Q=I),f_{s}$ as the key, and $f_{t}$ as the value $(K=f_{s}$ $V=f_{t})$ , CMM enhance features through an attention mechanism to obtain $I^\prime=\{I_1^{\prime},I_2^{\prime},...,I_n^{\prime}\}$ and map $I^\prime$ to the target domain class prototypes:
\begin{equation}
I' = \mathrm{softmax}\left(\frac{I f_s^T}{\sqrt{d}}\right) \cdot f_t + I
\label{eq:I_prime}
\end{equation}
\begin{equation}
P_{\mathrm{class}} = \mathrm{Linear}(I')
\label{eq:p_class}
\end{equation}
$I^{\prime}$furnishes a domain-invariant and semantically aligned
feature basis for the prompt, reducing the complexity of
learning cross-domain knowledge.
The prompt is supervised by cross-entropy loss to constrain its understanding of class semantics
CMM effectively maps features to a domain-invariant class
prototype space, standardizing semantic understanding across domains and mitigating source over-reliance. The conditional alignment loss $\mathcal{L}_{cal}$ supervises this process, ensuring the mapped features $I^\prime$ accurately reflect the pseudo label distributions.
\begin{table*}[htbp]
\centering
\footnotesize 
\setlength{\tabcolsep}{6pt} 
\renewcommand{\arraystretch}{0.9} 
\begin{tabular*}{\textwidth}{@{}c|>{\centering\arraybackslash}p{0.7cm}  ccccccccccccc @{}}
\toprule
\textbf{Method} & A$\to$C & A$\to$P & A$\to$R & C$\to$A & C$\to$P & C$\to$R & P$\to$A & P$\to$C & P$\to$R & R$\to$A & R$\to$C & R$\to$P & Avg\\
\midrule
TVT & 74.9 & 86.8 & 89.5 & 82.8 & 87.9 & 88.3 & 79.8 & 71.9 & 90.1 & 85.5 & 74.6 & 90.6 & 83.6 \\
SDAT & 69.1 & 86.6 & 88.9 & 81.9 & 86.2 & 88.0 & 81.0 & 66.7 & 89.7 & 86.2 & 72.1 & 91.9 & 82.4 \\
SSRT & \bfseries 75.2 & 89.0 & 91.1 & 85.1 & 88.3 & 89.9 & 85.0 & 74.2 & 91.2 & 85.7 & \bfseries 78.6 & 91.8 & 85.4 \\
Deit-based & 61.8 & 79.5 & 84.3 & 75.4 & 78.8 & 81.2 & 72.8 & 55.7 & 84.4 & 78.3 & 59.3 & 86.0 & 74.8 \\
CDTrans-Deit & 68.8 & 85.0 & 86.9 & 81.5 & 87.1 & 87.3 & 79.6 & 63.3 & 88.2 & 82.0 & 66.0 & 90.6 & 80.5 \\
\midrule
zero-shot CLIP & 67.9 & 89.0 & 89.4 & 82.4 & 89.0 & 89.4 & 82.4 & 67.6 & 89.4 & 82.4 & 67.6 & 89.0 & 82.1 \\
linear probe CLIP & 60.1 & 73.7 & 80.9 & 66.4 & 76.4 & 76.8 & 63.4 & 61.0 & 82.3 & 74.7 & 64.8 & 88.3 & 72.4 \\
CoOp & 70.0 & 90.8 & 90.9 & 83.2 & 90.9 & 89.2 & 82.0 & 71.8 & 90.5 & 83.8 & 71.5 & 92.0 & 83.9 \\
CoCoOp & 70.4 & 91.4 & 90.4 & 83.5 & \bfseries 91.8 & 90.3 & 83.4 & 67.0 & 91.0 & 83.4 & 71.2 & 91.7 & 84.1 \\
VP & 66.7 & 89.1 & 89.1 & 81.7 & 89.0 & 89.2 & 81.8 & 70.6 & 89.1 & 81.7 & 66.6 & 89.0 & 81.7 \\
VPT-shallow & 69.3 & 90.1 & 90.2 & 83.4 & 91.0 & 90.2 & 82.6 & 71.5 & 90.9 & 83.5 & 69.6 & 91.2 & 83.6 \\
VPT-deep & 71.6 & 89.9 & 90.3 & 82.8 & 91.0 & 89.7 & 82.0 & 72.4 & 90.3 & 84.6 & 71.7 & 91.6 & 83.9 \\
IVLP & 71.4 & \bfseries 91.7 & 90.5 & 83.6 & 90.2 & 89.3 & 82.2 & 71.6 & 90.4 & 84.1 & 72.1 & 92.0 & 84.2 \\
MaPLe & 72.2 & 91.6 & 90.3 & 82.6 & 90.9 & 89.8 & 82.4 & 70.7 & 90.1 & 85.1 & 72.0 & 92.1 & 84.2 \\
DAPL & 70.7 & 91.0 & 90.9 & 85.2 & 91.0 & 91.0 & 85.1 & 70.7 & 90.9 & 85.3 & 70.4 & 91.4 & 84.4 \\
PDA & 73.5 & 91.4 & 91.3 & 86.0 & 91.6 & \bfseries 91.5 & 86.0 & 73.5 & 91.7 & 86.4 & 73.0 & 92.4 & 85.7 \\
\bfseries C-DGPA (Ours) & 74.7 &  91.3 & \bfseries 91.6 & \bfseries 87.2 & 91.5 & 91.4 & \bfseries 86.9 & \bfseries 75.6 & \bfseries 91.9 & \bfseries 87.6 & 75.3 & \bfseries 93.2 & \bfseries 86.5 \\
\bottomrule
\end{tabular*}
\caption{Comparisons with prompt-tuning methods and SOTA methods using ViT-B/16 as the backbone network on the OfficeHome dataset. Bold indicates the best scores. }
\label{tab:full_comparison}
\end{table*}
\subsection{Synergistic Optimization and Inference}
The core strength of C-DGPA lies in the synergistic opti-
mization of its dual branches. Combined with the base
classification loss of CLIP $\mathcal{L}_\mathrm{cls}:$
\begin{equation}
\mathcal{L}_{cls}^S = -\frac{1}{N}\sum_{i=1}^N y_i^s \log\frac{\exp(\langle T_i, I^s \rangle/\tau)}{\sum_k \exp(\langle T_k, I^s \rangle/\tau)}
\label{eq:cls_S}
\end{equation}
\begin{equation}
\mathcal{L}_{cls}^T = -\frac{1}{M}\sum_{j=1}^M \hat{y}_j^t \log\frac{\exp(\langle T_j, I^t \rangle/\tau)}{\sum_k \exp(\langle T_k, I^t \rangle/\tau)}
\label{eq:cls_T}
\end{equation}
\begin{equation}
\mathcal{L}_{cls} = \mathcal{L}_{cls}^S + \mathcal{L}_{cls}^T
\label{eq:cls_total}
\end{equation}
The prompt parameters $\theta_p$ are updated end-to-end by
minimizing the combined loss:
\begin{equation}
\mathcal{L} = \mathcal{L}_{\mathrm{cls}} + \gamma_{\mathrm{cal}}\mathcal{L}_{\mathrm{cal}} + \gamma_{\mathrm{mal}}\mathcal{L}_{\mathrm{mal}}
\label{eq:total_loss}
\end{equation}
where $\gamma_{\mathrm{cal}}$ and $\gamma_{\mathrm{mal}}$are hyperparameters.
The Marginal Distribution Alignment Branch provides
domain-invariant features $I$,which are essential for the
CMM in the Conditional Distribution Alignment Branch to effectively align class-specific distribution. Conversely, the improved class discriminability facilitated by CMM refines the feature space, benefiting the domain discriminator.
This adaptive process progressively enhances both domain invariance and class separability. The prompt parameters are updated as follows:
\begin{equation}
\theta_p \leftarrow \theta_p - \eta \left( \nabla_{\theta_p}\mathcal{L}_{\mathrm{cls}} + \gamma_{\mathrm{cal}}\nabla_{\theta_p}\mathcal{L}_{\mathrm{cal}} + \gamma_{\mathrm{mal}}\nabla_{\theta_p}\mathcal{L}_{\mathrm{mal}} \right)
\label{eq:theta_update}
\end{equation}
\begin{table*}[htbp]
\centering
\footnotesize
\footnotesize 
\setlength{\tabcolsep}{6pt} 
\renewcommand{\arraystretch}{0.9} 
\begin{tabular}{@{}c|c|>{\centering\arraybackslash}p{0.5cm} cccccccccccc @{}}
\toprule
\textbf{Backbone} & \textbf{Method} & \textbf{plane} & \textbf{bicycle} & \textbf{bus} & \textbf{car} & \textbf{horse} & \textbf{knife} & \textbf{mcycl} & \textbf{person} & \textbf{plant} & \textbf{sktbrd} & \textbf{train} & \textbf{truck} & \textbf{Avg} \\
\midrule
\multirow{7}{*}{\rotatebox[origin=c]{-90}{ResNet101}}
& ERM & 55.1 & 53.3 & 61.9 & 59.1 & 80.6 & 17.9 & 79.7 & 31.2 & 81.0 & 26.5 & 73.5 & 8.5 & 52.4 \\
& DANN & 81.9 & 77.7 & 82.8 & 44.3 & 81.2 & 29.5 & 65.1 & 28.6 & 51.9 & 54.6 & 82.8 & 7.8 & 57.4 \\
& SDAT & 94.8 & 77.1 & 82.8 & 60.9 & 92.3 & 95.2 & 91.7 & 79.9 & \bfseries 89.9 & \bfseries 91.2 & 88.5 & 41.2 & 82.1 \\
& MCC & 88.1 & 80.3 & 80.5 & 71.5 & 90.1 & 93.2 & 85.0 & 71.6 & 89.4 & 73.8 & 85.0 & 36.9 & 78.8 \\
& MCD & 87.0 & 60.9 & 83.7 & 64.0 & 88.9 & 79.6 & 84.7 & 76.9 & 88.6 & 40.3 & 83.0 & 25.8 & 71.9 \\
& SHOT & 94.3 & \bfseries 88.5 & 80.1 & 57.3 & 93.1 & \bfseries 94.9 & 80.7 & \bfseries 80.3 & 91.5 & 89.1 & 86.3 & 58.2 & 82.9 \\
& C-DGPA (Ours) & \bfseries 98.2 & 84.5 & \bfseries 91.4 & \bfseries 72.6 & \bfseries 97.5 & 93.3 & \bfseries 94.3 & 79.3 & 86.8 & 86.1 & \bfseries 91.1 & \bfseries 63.5 & \bfseries 86.6 \\
\midrule
\multirow{9}{*}{\rotatebox[origin=c]{-90}{ViT-B/16}}
& CLIP & 99.2 & 92.2 & 93.5 & 76.7 & 98.3 & 90.4 & 94.6 & 83.6 & 85.4 & 96.1 & 94.3 & 62.7 & 88.9 \\
& VPT & 98.7 & 78.2 & \bfseries 96.0 & 72.8 & 98.8 & 70.5 & \bfseries 98.2 & 82.5 & 87.4 & 93.1 & 94.3 & 54.6 & 85.4 \\
& CoOp & 98.7 & 88.8 & 87.2 & 69.7 & 99.0 & 71.5 & 96.3 & 53.9 & 91.5 & 96.3 & 95.8 & 35.7 & 82.0 \\
& CoCoOp & 99.1 & 92.4 & 92.0 & 71.7 & 99.1 & 73.4 & 95.8 & 44.5 & 90.3 & 95.6 & \bfseries 96.0 & 52.8 & 83.6 \\
& IVLP & 98.2 & 71.2 & 82.6 & 79.9 & 97.3 & 68.3 & \bfseries 98.2 & 59.0 & \bfseries 90.5 & 93.4 & 95.7 & 36.6 & 80.9 \\
& MaPLe & 98.4 & 83.4 & 88.8 & 67.8 & 98.8 & 75.2 & 95.7 & 77.7 & 81.7 & 95.4 & 95.6 & 40.4 & 83.2 \\
& DAPL & 98.9 & 92.6 & 93.1 & 77.7 & 98.6 & 91.1 & 94.4 & 83.5 & 87.5 & 95.9 & 93.7 & 63.7 & 89.2 \\
& PDA & \bfseries 99.2 & 91.1 & 91.9 & 77.1 & 98.4 & 93.6 & 95.1 & 84.9 & 87.2 & \bfseries 97.3 & 95.3 & \bfseries 65.3 & 89.7 \\
& C-DGPA (Ours) & \bfseries 99.2 & \bfseries 94.3 & 94.0 & \bfseries 77.6 & \bfseries 99.2 & \bfseries 94.3 & 94.4 & \bfseries 85.6 & 88.8 & 95.4 & 94.8 & 65.1 & \bfseries90.2 \\
\bottomrule
\end{tabular}%
\caption{Comparisons with prompt-tuning methods and SOTA methods using ViT-B/16 and ResNet101 as backbone networks on the VisDA-17 dataset. Bold indicates the best scores.}
\label{tab:performance_comparison_categories}
\end{table*}
\section{Experiments}
\subsection{Experimental Setting}
\subsubsection{Datasets:}Experiments were conducted on commonly used, high-quality unsupervised domain adaptation benchmark datasets, including Office31\cite{saenko2010adapting}, OfficeHome\cite{venkateswara2017deep}, and VisDA-2017\cite{peng2018visda}.
\subsubsection{Baselines:} The work compared C-DGPA with state-of-the-art UDA methods (both fine-tuning methods and SOTA methods). For comparisons using the ViT-B/16\cite{dosovitskiy2020image} backbone, the models included TVT\cite{yang2023tvt}, SDAT\cite{rangwani2022closer}, SSRT\cite{sun2022safe}, Deit-based\cite{touvron2021training}, CDTrans-Deit\cite{xu2021cdtrans}, CLIP\cite{hu2024reclip,lai2024empowering}, CoOp\cite{zhou2022learning}, CoCoOp\cite{zhou2022conditional}, VP\cite{bahng2022exploring}, VPT\cite{jia2022visual}, IVLP\cite{khattak2023maple}, MaPle\cite{khattak2023maple}, DAP\cite{ge2023domain}, and PDA\cite{bai2024prompt}. For comparisons using the ResNet101\cite{dosovitskiy2020image} backbone, the models included ERM, DANN\cite{ganin2016domain}, MCD\cite{saito2018maximum}, MCC\cite{zhang2019bridging}, SDAT \cite{rangwani2022closer}, and SHOT\cite{liang2020we}.
\subsubsection{Experimental Setup:}In terms of experimental configurations, the work selected ResNet101 and ViT-B/16 as the backbone networks. For ResNet-based backbones, text prompts were used for prompt design, while multimodal prompts were employed for ViT-based backbones. The parameters in the CLIP encoder were kept fixed. The work used the SGD optimizer to train the prompts for 10 epochs on the VisDA-2017 dataset and 30 epochs on the Office31 and OfficeHome datasets, with a batch size of 32. The initial learning rate was set to 0.003 and was decreased according to the cosine annealing schedule. Additionally, the context length for C-DGPA was set to 16.
\subsection{Comparisons on the OfficeHome Dataset}
As shown in Table 2, C-DGPA achieves an average accuracy of 86.5\% on the OfficeHome dataset, surpassing all comparison methods. This represents a significant improvement of 4.4\% compared to zero-shot CLIP (82.1\%). Specifically, in the subtask "Real-World to Clipart" (R→C), C-DGPA attains an accuracy of 75.3\%, outperforming PDA (73.0\%), MaPLe (72.0\%), and VPT-deep (71.7\%) by 2.3\%, 3.3\%, and 3.6\%, respectively. Additionally, in tasks with larger cross-domain differences, such as "Product to Clipart" (P→C), C-DGPA demonstrates particularly strong performance, with accuracy leading SDAT and SSRT by 8.9\% and 1.4\%, respectively. Notably, IVLP and DAPL, which are based on multimodal prompts, lag behind C-DGPA by 2.3\% and 2.1\% in average performance on the OfficeHome dataset, further validating the effectiveness of aligning both marginal and conditional distribution simultaneously.
\subsection{Comparisons on the VisDA-2017 Dataset}
As shown in Table 3, on the VisDA-2017 dataset, C-DGPA establishes a new state-of-the-art by obtaining 90.2 \% accuracy with the ViT-B/16 backbone, surpassing prompt-tuning leaders PDA (89.7 \%), DAPL (89.2 \%), MaPLe (83.2 \%) and CoOp (82.0 \%) by clear margins of 0.5–8.2 percentage points , while with ResNet101 it reaches 86.6 \%, comfortably exceeding MCD (71.9 \%), MCC (78.8 \%) and even SDAT (82.1 \%); notably, SHOT—also built on ResNet101—stops at 82.9 \%, further attesting that C-DGPA’s dual alignment of marginal and conditional distributions delivers robust, architecture-agnostic gains. 
\subsection{Comparisons on the Office31 Dataset}
As shown in Table 4, C-DGPA achieved the best performance on the Office31 dataset, with an average accuracy of 91.8\%, representing a significant improvement of 14.3\% over zero-shot CLIP. In most tasks, C-DGPA outperformed other methods. For example, in the A→D task, C-DGPA's accuracy exceeded that of PDA, MaPLe, and VPT-deep by 2.4\%, 6.7\%, and 4.0\%, respectively. C-DGPA also achieved the best performance in the A→W, D→W, and W→D tasks. Additionally, The work noted that although CoCoOp is built upon CoOp, its average performance on the Office31 dataset was 0.5\% lower than that of CoOp.
\begin{table}[htbp]
\centering
\footnotesize
\setlength{\tabcolsep}{4pt}

\begin{tabular}{@{}c|>{\centering\arraybackslash}p{0.7cm} ccccccc @{}}
\toprule
\textbf{Method} & \textbf{A-D} & \textbf{A-W} & \textbf{D-A} & \textbf{D-W} & \textbf{W-A} & \textbf{W-D} & \textbf{Avg} \\
\midrule
zero-shot CLIP & 77.7 & 75.8 & 79.0 & 75.8 & 79.0 & 77.7 & 77.5 \\
linear probe CLIP & 83.1 & 83.3 & 74.2 & 96.5 & 70.3 & 98.4 & 84.3 \\
CoOp & 88.5 & 88.5 & 82.0 & 96.1 & 82.4 & 99.0 & 89.4 \\
CoCoOp & 86.9 & 88.2 & 83.2 & 94.1 & \bfseries 82.8 & 98.0 & 88.9 \\
VP & 78.5 & 74.8 & 77.9 & 77.5 & 77.8 & 79.7 & 77.4 \\
VPT-shallow & 83.5 & 83.8 & 77.5 & 88.6 & 80.9 & 91.2 & 84.2 \\
VPT-deep & 89.6 & 86.5 & 81.9 & 96.5 & \bfseries 82.8 & 99.2 & 89.4 \\
IVLP & 85.7 & 89.2 & 81.9 & 98.4 & 80.3 & 99.2 & 89.1 \\
MaPLe & 86.9 & 88.6 & 83.0 & 97.7 & 82.0 & 99.4 & 89.6 \\
DAPL & 81.7 & 80.3 & 81.2 & 81.8 & 81.0 & 81.3 & 81.2 \\
PDA & 91.2 & 92.1 & \bfseries 83.5 & 98.1 & 82.5 & \bfseries 99.8 & 91.2 \\
C-DGPA (Ours) & \bfseries 93.6 & \bfseries 93.7 & 82.6 & \bfseries 98.6 & 82.4 & \bfseries 99.8 & \bfseries 91.8 \\
\bottomrule
\end{tabular}%
\caption{Comparison of prompt-tuning methods using ViT-B/16 as the backbone network on the Office31 dataset. Bold indicates the best scores.}
\label{tab:performance_comparison}
\end{table}
\subsection{Ablation Study}
\subsubsection{Ablation Study on Dual-Branch Alignment Strategy:}Table 5 demonstrates that combining $\mathcal{L}_{mal}$ and $\mathcal{L}_{cal}$ yields optimal performance: Office31/ViT (91.8\%), OfficeHome/ViT (86.5\%), VisDA-2017/ViT (90.2\%), and VisDA-2017/Res (86.6\%). Using only $\mathcal{L}_{mal}$ slightly reduces accuracy (e.g., -0.3\% on OfficeHome/ViT), while $\mathcal{L}_{cal}$ alone leads to a larger drop (-0.8\%). On VisDA-2017, dual-loss optimization provides stable gains (+0.2–0.3\%), confirming their complementary effect. The same trend holds for ResNet, proving architecture-agnostic universality.
\begin{table}[htbp]
\centering
\footnotesize
  \setlength{\tabcolsep}{1pt}

\begin{tabular}{cc|ccccc}
\toprule
\bfseries \textbf{$\mathcal{L}_{mal}$} & \textbf{$\mathcal{L}_{cal}$} & \textbf{Office31/ViT} & \textbf{OfficeHome/ViT} & \textbf{Visda2017/ViT} & \textbf{Visda2017/Res} \\
\midrule
\ding{51} & \ding{51} & \bfseries 91.8 & \bfseries 86.5 & \bfseries 90.2 & \bfseries 86.6 \\
\ding{55} & \ding{51} & 91.3 & 86.2 & 90.0 & 86.4 \\
\ding{51} & \ding{55} & 91.1 & 85.7 & 89.9 & 86.1 \\
\bottomrule
\end{tabular}%

\caption{The impact of different branches (Marginal + Conditional) on domain adaptation. }
\label{tab:ablation_study}
\end{table}
\subsection{Analysis of Key Hyperparameters}
\subsubsection{Analysis Weight Of Marginal And Conditional Alignment Loss:}As shown in Table 6, the hyperparameter study reveals critical tradeoffs for marginal alignment loss weight ($\gamma_{mal}$) and conditional alignment loss weight ($\gamma_{cal}$). An optimal $\gamma_{mal}=0.01$ maximizes average accuracy (86.5\% on OfficeHome), with strong performance in both A domain (87.2\%) and P domain (92.0\%). However, excessive marginal alignment weighting ($\gamma_{mal}=2$) over-suppresses domain-specific features, reducing discriminability and causing a 0.8\% average accuracy drop (85.7\%). For conditional alignment, $\gamma_{cal}=1$ achieves the best equilibrium (C domain: 75.2\%), while increasing to $\gamma_{cal}=2$ boosts P domain accuracy to 92.0\% but destabilizes other domains. Conversely, reducing to $\gamma_{cal}=0.1$ weakens class alignment, lowering C domain accuracy by 1.1\%.
\begin{table}[htbp]
\centering
\scriptsize 
\setlength{\tabcolsep}{1.5pt} 
\renewcommand{\arraystretch}{0.75} 
\begin{minipage}[t]{\linewidth}
\centering
\begin{minipage}[t]{0.49\linewidth}
\centering
\begin{tabular}{@{}r|rrrrr@{}}
\toprule
{\textbf{$\gamma_{mal}$}} & \textbf{A} & \textbf{C} & \textbf{P} & \textbf{R} & \textbf{Avg} \\
\midrule
2 & 86.6 & 73.8 & 91.2 & 91.1 & 85.7 \\
1 & 86.7 & 73.8 & 91.1 & 91.2 & 85.7 \\
0.1 & 87.0 & 74.7 & 91.8 & 91.6 & 86.3 \\
\bfseries 0.01 & \bfseries 87.2 & \bfseries 75.2 & \bfseries 92.0 & \bfseries 91.6 & \bfseries 86.5 \\
\bottomrule
\end{tabular}
\caption*{(a) The impact of $\gamma_{mal}$ \\ on OfficeHome.}
\end{minipage}
\hfill
\begin{minipage}[t]{0.49\linewidth}
\centering
\begin{tabular}{@{}r|rrrrr@{}}
\toprule
{\textbf{$\gamma_{cal}$}} & \textbf{A} & \textbf{C} & \textbf{P} & \textbf{R} & \textbf{Avg} \\
\midrule

2 & 86.9 & 74.2 & \bfseries 92.0 & \bfseries 91.7 & 86.2 \\
\bfseries 1 & \bfseries 87.2 & \bfseries 75.2 & \bfseries 92.0 & 91.6 & \bfseries 86.5 \\
0.1 & \bfseries 87.2 & 74.1 & 91.8 & 91.4 & 86.1 \\
0.01 & 86.0 & 73.4 & 91.6 & 91.2 & 85.6 \\
\bottomrule
\end{tabular}
\caption*{(b) The impact of $\gamma_{cal}$ \\ on OfficeHome.}
\end{minipage}
\end{minipage}
\caption{Performance comparison under different $\gamma$ settings}
\end{table}
\begin{figure}[htbp]
    \centering
    \begin{subfigure}[b]{0.49\linewidth}
        \includegraphics[width=\linewidth]{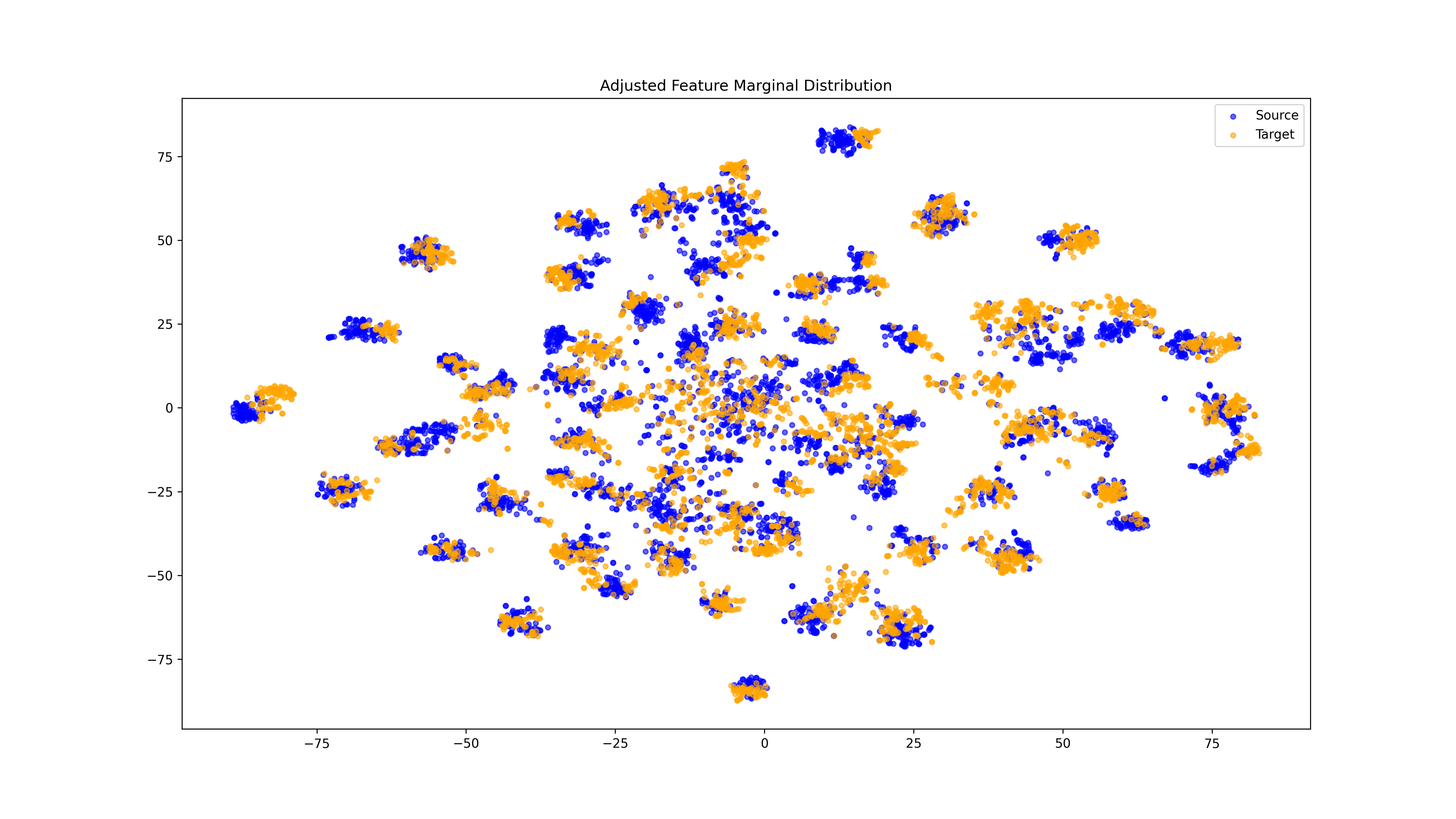}
        \caption{C-DGPA}
        \label{fig:left_cond}
    \end{subfigure}
    \hfill
    \begin{subfigure}[b]{0.49\linewidth}
        \includegraphics[width=\linewidth]{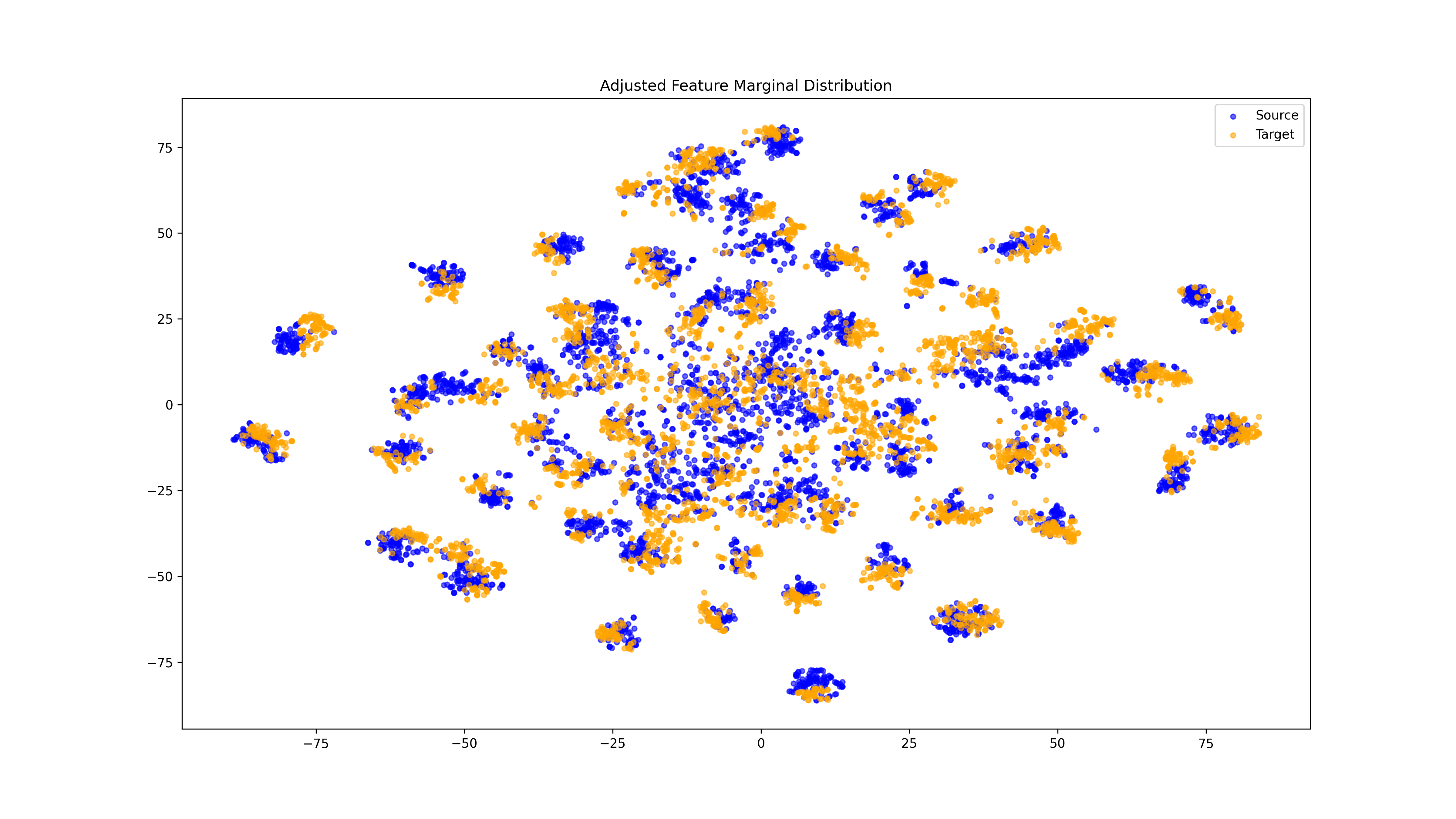}
        \caption{PDA}
        \label{fig:right_cond}
    \end{subfigure}
    \caption{t-SNE visualization of source and target domain features, highlighting marginal distribution using C-DGPA and PDA (OfficeHome dataset, R→P). }
    \label{fig:both}
\end{figure}
\begin{figure}[htbp]
    \centering
    \begin{subfigure}[b]{0.48\linewidth}
        \includegraphics[width=\linewidth]{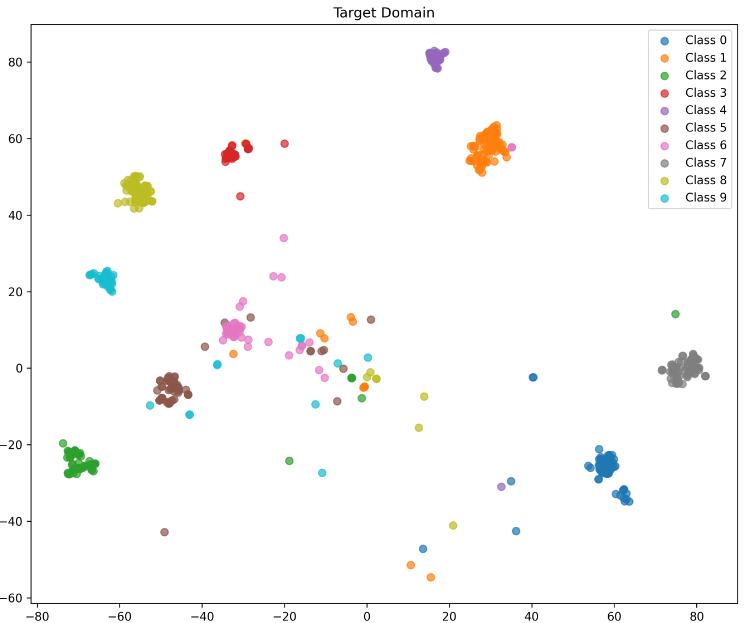}
        \caption{C-DGPA}
        \label{fig:left}
    \end{subfigure}
    \hfill
    \begin{subfigure}[b]{0.48\linewidth}
        \includegraphics[width=\linewidth]{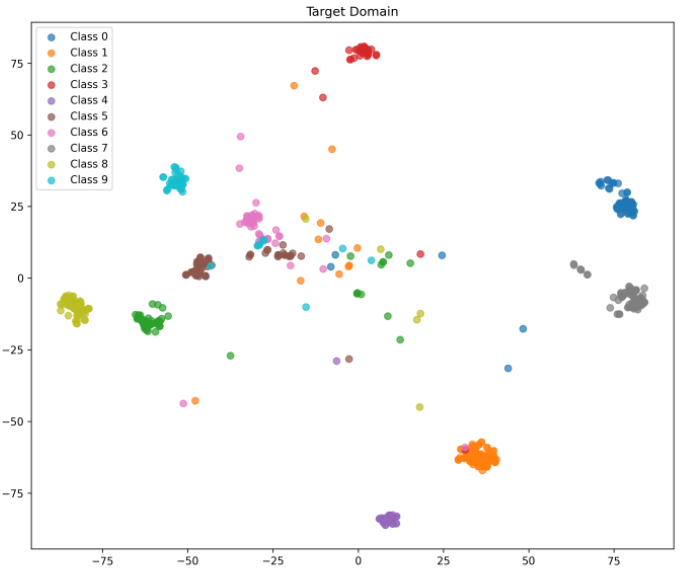}
        \caption{PDA}
        \label{fig:right}
    \end{subfigure}
    \caption{t-SNE visualization of target domain features, highlighting conditional  distribution using C-DGPA and PDA (OfficeHome dataset, R→P).}
    \label{fig:both}
\end{figure}
\subsection{Visualization Analysis}
The t-SNE visualization (Figure 2) shows that PDA (only marginal alignment) results in partial domain overlap but significant same-class center offset. Target domain features are dispersed with blurred boundaries, indicating poor conditional distribution alignment. In contrast (Figure 3), C-DGPA (joint marginal + conditional alignment) produces highly overlapping target features with compact intra-class structure and clear inter-class boundaries. This demonstrates the effectiveness of joint alignment in mitigating class prototype shift.
\begin{figure}
    \centering
    \includegraphics[width=0.8\linewidth]{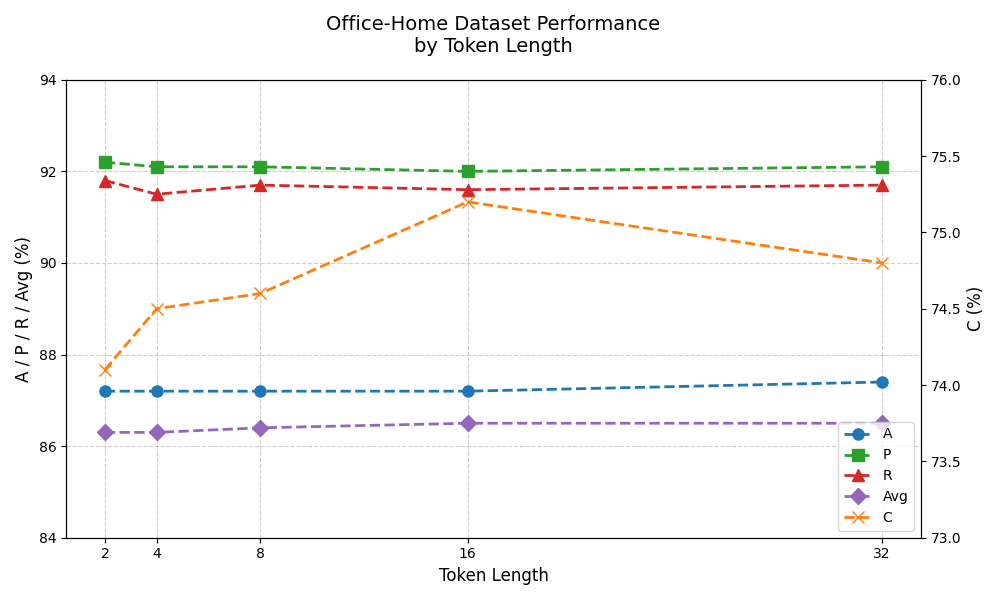}
    \caption{he impact of different token lengths on accuracy (OfficeHome dataset) }
    \label{fig:token_len}
\end{figure}

\subsubsection{The Impact of Token Length on Accuracy:}Figure 4 shows token length's impact on OfficeHome accuracy. Accuracy peaks at 86.5\% with 16 tokens, rising initially before stabilizing. Shorter tokens ($\le$ 4) limit semantic capacity, reducing C domain performance (74.1\%-74.5\%). Longer tokens ($\ge$ 16) marginally improve A domain accuracy (87.4\%) but increase computational costs with diminishing returns. A length of 16 optimally balances semantic expression and efficiency.
\section{Conclusion}
The work introduces C-DGPA, a class-centric dual-alignment framework that adapts vision–language models to unsupervised domain adaptation by tackling the critical limitation of prior prompt-tuning methods: their exclusive focus on marginal distribution.  C-DGPA resolves the joint discrepancy—both marginal and conditional—via a synergistic two-branch architecture, yielding new state-of-the-art results on OfficeHome, Office31 and VisDA-2017. The marginal branch employs a dynamic adversarial minimax game with gradient reversal to align cross-domain feature marginals, while the conditional branch leverages a Class Mapping Mechanism that aligns class prototypes and mitigates semantic drift. Jointly optimized, the branches reciprocally enhance domain invariance and class discrimination.  Future work will focus on extending C-DGPA to more complex scenarios, such as source-free domain adaptation, multi-target domain adaptation, and domain generalization, to further improve its applicability and effectiveness in a broader range of tasks.
\section{Acknowledgments}
To maintain anonymity, this section will remain blank for now. If it passes the review, We will provide the necessary additions later.

\bibliography{aaai2026}

\end{document}